\documentclass[conference]{IEEEtran}

\IEEEoverridecommandlockouts
\usepackage[noadjust]{cite}
\usepackage[inline]{enumitem}
\usepackage{amsmath,amssymb,amsfonts}
\usepackage{gensymb}
\usepackage{algorithmic}
\algsetup{linenosize=\tiny}
\usepackage{graphicx}
\usepackage{booktabs}
\usepackage{textcomp}
\usepackage{hyperref}
\usepackage[table]{xcolor}
\definecolor{Green}{rgb}{0.0, 1.0, 0.0}
\definecolor{Cyan}{rgb}{0.0, 1.0, 1.0}
\definecolor{LightRed}{rgb}{1.0, 0.5, 0.5}
\definecolor{Orange}{rgb}{1.0, 0.65, 0.0}
\definecolor{Purple}{rgb}{0.5, 0.0, 0.5}
\definecolor{Yellow}{rgb}{1.0, 1.0, 0.0}
\usepackage[linesnumbered,ruled]{algorithm2e}
\definecolor{bleudefrance}{rgb}{0.19, 0.55, 0.91}

\ifCLASSOPTIONcompsoc
\usepackage[caption=false,font=normalsize,labelfon
t=sf,textfont=sf]{subfig}
\else
\usepackage[caption=false,font=footnotesize]{subfi
g}
\fi
\newcommand{\norm}[1]{\left\lVert #1 \right\rVert}
\def\BibTeX{{\rm B\kern-.05em{\sc i\kern-.025em b}\kern-.08em
    T\kern-.1667em\lower.7ex\hbox{E}\kern-.125emX}}
  
\begin{document}

\title{Optimizing Cooperative Multi-Object Tracking using Graph Signal Processing 
\thanks{This work has received funding from the EU’s Horizon Europe research and innovation programme in the frame of the AutoTRUST project “Autonomous self-adaptive services for TRansformational personalized inclUsivenesS and resilience in mobility” under the Grant Agreement No 101148123.
}
}
\author{
\IEEEauthorblockN{Maria Damanaki$^{1,3}$, Nikos Piperigkos$^{1,2}$, Alexandros Gkillas$^{1,2}$, Aris S. Lalos$^{1,2}$}
\IEEEauthorblockA{$^1$Industrial Systems Institute, Athena Research Center, Patras Science Park, Greece\\
$^2$AviSense.AI, Patras Science Park, Greece, $^3$ Dpt. of Informatics \& Telecom., University of Ioannina, Arta, Greece\\
Emails: mdamanaki@isi.gr, \{piperigkos, gkillas\}@avisense.ai, lalos@athenarc.gr 
}
}
\maketitle
\begin{abstract}
Multi-Object Tracking (MOT) plays a crucial role in autonomous driving systems, as it lays the foundations for advanced perception and precise path planning modules. Nonetheless, single agent based MOT lacks in sensing surroundings due to occlusions, sensors failures, etc. Hence, the integration of multi-agent information is essential for comprehensive understanding of the environment. This paper proposes a novel Cooperative MOT framework for tracking objects in 3D LiDAR scene by formulating and solving a graph topology-aware optimization problem so as to fuse information coming from multiple vehicles. By exploiting a fully connected graph topology defined by the detected bounding boxes, we employ the Graph Laplacian Processing optimization technique to smooth the position error of bounding boxes and effectively combine them. In that manner, we reveal and leverage inherent coherences of diverse multi-agent detections, and associate the refined bounding boxes to tracked objects at two stages, optimizing localization and tracking accuracies. An extensive evaluation study has been conducted, using the real-world V2V4Real dataset, where the proposed method significantly outperforms the baseline frameworks, including the state-of-the-art deep-learning DMSTrack and V2V4Real, in various testing sequences.
\end{abstract}

\section{Introduction}
\label{intro}
The rapid growth of autonomous driving technology necessitates robust object detection and recognition capabilities to ensure safe and accurate operations in dynamic environments. Multi-Object Tracking (MOT) is a crucial functionality enabling the continuous identification and localization of objects in the scene \cite{10608725,10204123,9352500,Yin_2021_CVPR,9341164}. However, the performance of MOT is prone to degradation under partially or fully occluded structures, challenging weather conditions (e.g. rain, fog, or snow), or sensor failures. The exchange of environmental information regarding the surrounding context among Connected and Automated Vehicles (CAVs) significantly enhances situational awareness \cite{9537608}.
Hence, Cooperative Object Detection (COD) and Cooperative Multi-Object Tracking (CoMOT) are essential to enable objects detection, localization and tracking throughout the entire network of agents engaged in Vehicle-to-Vehicle (V2V) communication \cite{chiu2024probabilistic, 10587163,10203124,9812038,8653482}. 
Specifically, CoMOT enhances the robustness and efficiency of MOT by utilizing 3D detections obtained from 3D detectors deployed on each CAV. All the 3D bounding boxes are projected to a Global Coordinate System and represent the objects’ positions of all connected agents at each time step. Thereafter, detections and trajectories with specific ids are associated forming the core of the tracking framework. 

Thus, CoMOT constitutes a critical component of situational awareness while the relevant literature broadly classified into \textbf{early}, \textbf{inter} and \textbf{joint cooperative tracking} schemes, based on how multi-vehicle information is leveraged for object tracking. 
In \textbf{early cooperative tracking} scheme, cooperative detections are integrated in advance and correlated with trajectories in the association module, as proposed in V2V4Real \cite{10203124}. In \cite{10430224}, the detections' uncertainty is quantified and a second association module is introduced to handle unmatched detections and trajectories. However, CoMOT performance deteriorates with an increased number of observations in dynamic surroundings.
In the \textbf{inter-cooperative} paradigm, multi-agent information is fused into the association module alongside with trajectories as proposed in \cite{10588576} which prioritizes ego-vehicle detections and performs sequential associations with trajectories. DMSTrack \cite{chiu2024probabilistic} is a similar deep-learning CoMOT framework, where a Deep Neural Network (DNN) is trained to estimate the measurement uncertainty. However, prioritizing ego-vehicle as a primary source of information may exacerbate the risk of false associations.
Lastly, in the \textbf{joint cooperative-tracking} scheme, detection and tracking modules utilize shared perception data in parallel. In \cite{10148929}, historical information of objects is leveraged, and heterogeneous detection and tracking features are fused to infer 3D bounding boxes. Nonetheless, reliance on historical data may cause system's inability to recognize new objects, while multi-agent fusion increases the number of bounding boxes, requiring excessive computational resources for effective tracking in real time.

To overcome the limitations of the aforementioned schemes, a novel \textbf{early cooperative tracking} method is proposed, exploiting and fusing multi-vehicle detections in a compact and graph topology aware optimization manner. More specifically, a fully connected graph is formulated based on the detected bounding boxes from multiple agents. Afterwards, the centroids of the 3D detections are refined, applying Graph Laplacian operator, a Graph Signal Processing tool which exploits the spatial coherences of the geometry defined by the diverse detections of the multiple agents. Subsequently, the refined detections are associated with existing trajectories at two stages, depending on how we use the multi-agent information in the proposed Graph Processing framework. The accuracy of CoMOT is further increased and improved employing the well-known linear Kalman filter on each bounding box. 

Therefore, the main contributions of this study can be summarized as follows: 
\begin{itemize}
\item A fully connected graph topology based on multi-agent detections is formulated, facilitating the fusion of the detected bounding boxes provided by different CAVs. This is achieved using Graph Signal Processing, exploiting diverse but overlapped detected bounding boxes.
\item Bounding boxes are addressed in a unified and compact manner, instead of successively processing detections from each cooperating agent as in the state-of-the-art \cite{chiu2024probabilistic}, providing a framework that can be easily extended for efficiently processing detections from multiple CAVs. Furthermore, associating refined bounding boxes with tracks, in a two-stage process, yields significant improvements in the overall tracking performance. 
\item An extensive experimental evaluation has been conducted using the state-of-the-art V2V4Real dataset \cite{10203124}, the first real-world 3D tracking benchmark dataset for V2V perception. The results of the proposed approach demonstrates significant improvements (up to \textbf{19.32\%}) with respect to two state-of-the-art CoMOT methods, employing well-known MOT tracking metrics. 
\end{itemize}

\section{System Model and Preliminaries}
\label{preliminaries}
This Section introduces the system model, and preliminaries related to the association techniques as well as the Graph Signal Processing framework of Graph Laplacian Processing.

\subsection{System Model}
\label{systemmodel}
Consider a tracking-by-detection paradigm, since it is widely employed in MOT literature, as the foundation of our \textbf{early cooperative tracking} framework. Each agent $i$ obtains $M_i$ 3D bounding boxes at time $t$ from the output of a 3D detector on LiDAR data. Each 3D bounding box is described by $\boldsymbol{x_{D}^{(i,m,t)}} =[ x_{i,m} \ y_{i,m} \ z_{i,m} \ \theta_{i,m} \ h_{i,m} \ w_{i,m} 
\ l_{i,m}]^T \in \mathbb{R}^7$ with $m=0,1, \ldots M_i$, where $x_{i,m},y_{i,m},z_{i,m}$ is the centroid of the bounding box, $\theta_{i,m}$ the angle around the z-axis, and $h_{i,m},w_{i,m},l_{i,m}$ the height, width, and length. Hence, the set of detections of agent $i$ at time instance $t$ is described by $\mathcal{D}_i^t = \{\boldsymbol{x_{D}^{(i,1,t)}}, \boldsymbol{x_{D}^{(i,2,t)}}, \ldots \boldsymbol{x_{D}^{(i,M_i,t)}}\}\in \mathbb{R}^{M_{i}\times 7}$.
Furthermore, at time instant $t$, the state of tracked object $r$ is defined by $\boldsymbol{x_{T}^{(r,t)}}=[x_r \ y_r \ z_r 
\ \theta_r \ h_r \ w_r \ l_r \ u_{x_r} \ u_{y_r} \ u_{z_r}]^T  \in \mathbb{R}^{10}$, where $x_r,y_r,z_r$ represent its centroid, $\theta_r$ the angle around z-axes, $h_r,w_r,l_r$ the height, width, length, and $u_{x_r}, u_{y_r}, u_{z_r}$ the 3D linear velocity, respectively. Note that the tracked object is the expected outcome of the cooperative tracking, therefore it is not directly related with a specific agent. Based on that, we will employ a state transition model, as well as a measurement model, in order to perform linear Kalman Filtering: 
\begin{itemize} 
\item State transition model: 
\begin{align}
    \begin{split}
    \label{CV}
    \boldsymbol{x_{T}^{(r,t)}} = f(\boldsymbol{x_{T}^{(r,t-1)}},\boldsymbol{w_{T}^{(r,t)}}), \boldsymbol{w_{T}^{(r,t)}} \sim \mathcal{G}(0, \boldsymbol{\Sigma_{w}})
    \end{split}
\end{align}
\end{itemize}
\begin{itemize}
    \item Measurement model:
    \begin{align}
    \label{self_positioning}
        \begin{split} \boldsymbol{z_{T}^{(r,t)}} = \boldsymbol{x_{T}^{(r,t)}} + \boldsymbol{n_p^{(r,t)}},  \boldsymbol{n_p^{(r,t)}} \sim \mathcal{G}(0, \boldsymbol{\Sigma_{n}})
        \end{split}
    \end{align}
\end{itemize}
The state transition function $f(\cdot)$ can be defined using constant velocity model. Both models are degraded by additive white Gaussian Noise $\boldsymbol{w_{T}^{(r,t)}}$, and $\boldsymbol{n_p^{(r,t)}}$, respectively. According to linear Kalman Filter, each track's state can be defined by the prediction and update equations as follows:
\begin{itemize}
    \item State Prediction:
    \begin{align}
        \boldsymbol{\bar{x}_{T}^{(r,t)}} &= \boldsymbol{F}\boldsymbol{\hat{x}_{T}^{(r,t-1)}} + \boldsymbol{w_{T}^{(r,t)}} \label{eq:prediction_sta}\\
        \boldsymbol{{\bar{P}^{(r,t)}}} &= \boldsymbol{F\hat{P}^{(r,t-1)}F^{T} + Q^{(r,t)}} \label{eq:prediction_P}
    \end{align}
    \item State Update:
    \begin{align}
    \label{State Update}
        \boldsymbol{K^{(r,t)}} &=\boldsymbol{ {\bar{P}^{(r,t)}}H^{T} [H{\bar{P}^{(r,t)}}H^{T} + R^{(r,t)}]^{-1}} \\
        \boldsymbol{\hat{x}_{T}^{(r,t)}} &= \boldsymbol{\bar{x}_{T}^{(r,t)}} + \boldsymbol{K^{(r,t)}[\boldsymbol{z_{T}^{(r,t)}} - H \boldsymbol{\bar{x}_{T}^{(r,t)}}] }\label{eq:upd_x} \\
        \boldsymbol{\hat{P}^{(r,t)}} &= \boldsymbol{ {\bar{P}^{(r,t)}} - K^{(r,t)}H{\bar{P}^{(r,t)}}} \label{eq:upd_P}
    \end{align}
\end{itemize}
where $\boldsymbol{F}\in \mathbb{R}^{10\times10}$ is the transition matrix, $\boldsymbol{Q^{(r,t)}}\in \mathbb{R}^{10\times10}$ the process noise covariance matrix, $\boldsymbol{R^{(r,t)}}\in \mathbb{R}^{10\times10}$ the measurement noise covariance matrix, $\boldsymbol{H}\in \mathbb{R}^{7\times10}$ the measurement model matrix, $\boldsymbol{K^{(r,t)}}\in \mathbb{R}^{10\times7}$ the Kalman gain and $\boldsymbol{P^{(r,t)}}\in \mathbb{R}^{10\times10}$ the covariance matrix of the tracked object. 
Finally, set $\mathcal{T}^t = \{\boldsymbol{\hat{x}_{T}^{(1,t)}}, \boldsymbol{\hat{x}_{T}^{(2,t)}},...,\boldsymbol{\hat{x}_{T}^{(R_T,t)}}\}\in \mathbb{R}^{R_T\times 10}$ contains the state of tracked objects $r=1,2, \ldots R_T$, after performing Kalman Filtering. Those two sets of detected and tracked bounding boxes will act as the basis in order to formulate the proposed framework.

\subsection{Association and Tracks Management approaches}
\label{Tracking Management Module}
The Association and Tracks Management modules constitute the core components of a tracking framework handling tracks' states and their lifetimes. Specifically, the Association module employs the 3D Intersection over Union (3D IoU) as a similarity metric, and the Hungarian Algorithm (HA) as an association algorithm to correlate trajectories and detections. Upon successful association, the track's state is updated utilizing the corresponding detection $\boldsymbol{x_D^{(i,m,t)}}$ according to Eq.({\ref{eq:upd_x}},{\ref{eq:upd_P}}), where $\boldsymbol{z_{T}^{(r,t)}=x_{D}^{(i,m,t)}}$. Conversely, on unsuccessful association, the track's state retains its previous state without update. Upon on unmatched detections, new tracks are initialized with their attributes. The Tracks Management module is handling the lifetime of each track and establishes the tracking procedure after the association stages with detections \cite{9341164},\cite{9711021}. Specifically, a track is "confirmed", if it achieves successful associations over a predefined number of consecutive frames, denoted as $hits$. 
Conversely, a track is terminated and removed if it fails to associate for a predefined number of consecutive time steps, defined as its $age$.

\subsection{Graph Signal Processing Preliminaries}
The Graph Laplacian operator is a well-known Graph Signal Processing Tool \cite{sorkine2005laplacian} which recovers the absolute coordinates of graph vertices through differential coordinates and anchor points in a least-squares sense. The inherent geometry of the connectivity graph is captured by differential coordinates of the vertices, which correspond to the barycenter of each vertex's neighboring nodes. Additionally, anchors are utilized as complementary information for each vertex. Furthermore, the Laplacian matrix captures the connections between the vertices. 
More specifically, consider an undirected graph with $N^t$ nodes represented as $J^{t} = (\mathcal{V}^t, \mathcal{E}^t)$, where $\mathcal{V}^t$ and $\mathcal{E}^t$ denote the set of vertices and edges, respectively. The objective is to define and minimize the cost function $O(\boldsymbol{v^t})= \norm{\boldsymbol{L^tv^t-\delta^t}}^2_2$ in each spatial attribute $x,y,z$,  
where $\boldsymbol{v^t}=[v^{(1,t)} \  v^{(2,t)} ... \ v^{(N_t,t)}]^T \in \mathbb{R}^{N^t}$ is the vector of graph vertices, $\boldsymbol{L^t} \in \mathbb{R}^{N^{t} \times{N^{t}}}$ is the Laplacian matrix and $\boldsymbol{\delta^t} \in \mathbb{R}^{N^{t}}$ the vector of differential coordinates. Laplacian matrix is equal to $\boldsymbol{L^t=D^t-A^t}$, where  $\boldsymbol{D^t}$ , $\boldsymbol{A^{t}} \in \mathbb{R}^{{N^t}\times{N^t}}$ are the well-known degree and adjacency matrices. Furthermore, in each spatial attribute the differential vector is equal to $\boldsymbol{\delta^{t}} = [\delta^{(1,t)} \  \delta^{(2,t)} \ \hdots \ \delta^{(N_t,t)}]$,  
where $\delta^{(i,t)}= \sum_{i}^{{N^t}}({v^{(i,t)}}-{v^{(j,t)}})$ among all connected vertices.
However, due to the singular properties of Laplacian matrix, we have to extend it using the identity matrix, leading to the extended Laplacian matrix $\boldsymbol{\tilde{L}^{t}} \in \mathbb{R}^{2N^t\times N^t}$, and thus, $\boldsymbol{\delta^t}$ is restructured to the measurement vector $\boldsymbol{b}^{t} \in \mathbb{R}^{2N^t}$ consisting of not only differential coordinates but also the so-called anchor points, which contain any knowledge we have about each vertex. Previous quantities, as well as anchors vector $\boldsymbol{a^t}\in \mathbb{R}^{N^t}$ are defined as follows: 
\begin{align}  
\boldsymbol{\tilde{L}^{t}}&=\begin{bmatrix}
        \boldsymbol{D^{t}-A^{t}} \\
       \boldsymbol{I^t}
    \end{bmatrix}^T ,
    \label{eq:exteLap}
    \boldsymbol{b^{t}} = \begin{bmatrix}
        \boldsymbol{\delta^{t}} &
        \boldsymbol{a^t}
    \end{bmatrix}^T
\end{align}
\begin{align}
    \boldsymbol{a^{t}} &= 
    \begin{bmatrix}
      a^{(1,t)} &
      a^{(2,t)} & 
      \hdots &
      a^{(N^t,t)}  
    \end{bmatrix}^T
    \label{eq:anchors}
\end{align}
Employing linear least-squares minimization, the optimal solution of vertices' locations in each spatial attribute is:
\begin{align}
    \boldsymbol{v_{*}^t} &= (\boldsymbol{(\tilde{L}^{t})^T\tilde{L}^{t})^{-1}(\tilde{L}^{t})^{T}}\boldsymbol{b^t}, \ \ \ \boldsymbol{v_{*}^t} \in \mathbb{R}^{N^t}
    \label{eq:LSMsol}
\end{align}
\begin{figure}[htbp]
\centering
 \includegraphics[scale=0.4]{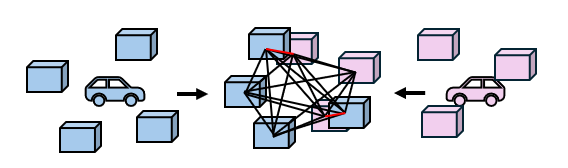}
    \caption{Fully connected graph of detections from two neighboring agents $i$ and $j$ at time {t}. Detections from agent $i$ (blue color) are interconnected among them, and also linked to detections from agent $j$ (pink color). Red and black edges indicate the overlapped information of the object shared between the agents and all the connections of the graph, respectively.}
  \label{fig:detections_graph}
\end{figure}

\section{Cooperative Multi-Object Tracking via Graph Topology Aware Optimization}
\subsection{Graph Laplacian Processing Technique}
The Graph Laplacian Processing technique actually facilitates the localization of nodes of a graph by leveraging inherent geometric structure defined by the differential coordinates and anchors. Based on that fact, two methods will be derived, introducing the Graph Laplacian CoMOT scheme which takes advantage of a fully connected graph among all multi-agent detections as illustrated in Fig.\ref{fig:detections_graph}. The one-stage of association approach serves as the initial step for the proposed two-stage association Graph Lap CoMOT method. Both of the presented approaches reduce spatial noise of vehicle's detections by smoothing the individual error related to the bounding box's 3D centroid. Therefore, each bounding box is refined and correlated with existing tracks, enabling more accurate trajectory estimation during the Kalman Filter update step.  
More specifically, differential coordinates and Laplacian matrix capture the connections among all detections, while anchors are utilized for the overlapped detections that represent objects simultaneously observed from different agents.  Although our methods are presented assuming two CAVs, following the standard methodology of \cite{chiu2024probabilistic}, they can easily be generalized to a larger number of connected vehicles. 

To be more specific, consider the undirected graph ${J}^{t} = (\mathcal{V}^t, \mathcal{E}^t)$, where the set of nodes  $\mathcal{V}^t=\{\mathcal{D}_i^{t},\mathcal{D}_j^{t}\}$ 
represents the detections from agent $i$ and $j$.
With respect to the set of edges $ \mathcal{E}^t$, the detections of agent $i$ are interconnected among them and also linked to the detections of agent $j$. Note that we focus only on estimating the 3D centroid of each bounding box. Therefore, in each spatial attribute $x,y,z$, the differential vector is equal to $\boldsymbol{\delta^t}= [\delta^{(1,t)} \ \delta^{(2,t)} \ \hdots \ \delta^{(N^t,t)}]^T \in \mathbb{R}^{N^t}$,
where $\delta^{(m,t)} = \sum_{n=0}^{M_j} \left( {x_{i,m} - {x_{j,n}}} \right) 
    + \sum_{n=0}^{M_i} \left( {x_{i,m} - {x_{i,n}}} \right)$ describes the scalar differential coordinate of detection $m$.
Additionally, the extended Laplacian matrix is defined by Eq.\ref{eq:exteLap} corresponding to a fully connected graph, anchors vector $\boldsymbol{a^t} \in \mathbb{R}^{N^t}$  will contain the scalar complimentary information of x-part of $\boldsymbol{x_D^{(i,m,t)}}$ as analytically explained in the following, while measurement vector $\boldsymbol{b^t}= [\boldsymbol{\delta^t} \ \boldsymbol{a^t}]^T \in \mathbb{R}^{2N^t}$. Similar equations are followed for $y$ and $z$ attributes. Thus, the refined bounding boxes, since we actually refine their 3D centroids, will be further used by the core modules of CoMOT in order to enhance the overall accuracy.

\subsection{All in One Stage (AOS) Graph Lap-CoMOT}
\label{Method 1}  
A simple CoMOT which utilizes a fully connected graph topology on multi-agent detections is \textbf{All in One Stage (AOS) Graph Lap-CoMOT}. More specifically, this method refines and fuses multi-vehicle bounding boxes through the Graph Laplacian framework and thereafter, associates them with existing tracks. To acquire additional information at time $t$ for forming the anchors vector $\boldsymbol{a^t}$, the 3D detections of the two agents $\mathcal{D}_i^t$ and $\mathcal{D}_j^t$ are associated through 3D IoU and HA. The set of $m_i$ overlapped detections and $u_i$ non-overlapped of the $i$ are described by $m\mathcal{D}_i^t \in \mathcal{R}^{m_i\times 7}$, and $u\mathcal{D}_i^t \in \mathcal{R}^{u_i\times 7}$, respectively. Similarly, for the agent $j$, $m\mathcal{D}_j^t \in \mathcal{R}^{m_j\times 7}$, and $u\mathcal{D}_j^t \in \mathcal{R}^{u_j\times 7}$.
The anchors vector $\boldsymbol{a^t}$ is formed for $x$ axis as:
\begin{align}
    \boldsymbol{a^{t}} &= 
    \begin{bmatrix}
    \boldsymbol{mx_{j,m}} & \boldsymbol{mx_{i,m}} & \boldsymbol{ux_{i,m}} & \boldsymbol{ux_{j,m}}     
    \end{bmatrix}^T
     \label{eq:anchorsAOS}
\end{align}
where $\boldsymbol{mx_{im}}=[x_{i,1}\ x_{i,2}\ ...\ x_{i,m_{i}}]^T \in \mathbb{R}^{m_i}$, $\boldsymbol{mx_{jm}}=[ x_{j,1}\ x_{j,2}\ ... \ x_{j,m_{j}}]^T \in \mathbb{R}^{m_j}$ are the successfully associated detections of the agent $i$, $j$ and $\boldsymbol{ux_{im}}=[x_{i,1}\ x_{i,2}\ ... \  x_{i,u_{i}}]^T \in \mathbb{R}^{u_i}$, $\boldsymbol{ux_{jm}}=[x_{j,1} \ x_{j,2} \ ... \ x_{j,u_{j}}]^T \in \mathbb{R}^{u_j}$ the unmatched bounding boxes of $i$ and $j$, respectively. 
Therefore, the least-squares minimization of Eq.\ref{eq:LSMsol} at time $t$ is noted as $\boldsymbol{G^{t}} \in \mathbb{R}^{N^t}$ and describes the refined and fused $x$ attribute of the detections. Similar equations are followed for the $y,z$ attributes.
The set of refined detections which consist all the attributes on each 3D bounding box [\ref{systemmodel}] is $\mathcal{G}^t \in \mathbb{R}^{N^t\times 7}$. 
Then, the optimized detections $\mathcal{G}^t$ are associated with the existing tracks $\mathcal{T}^t$ in terms of 3D IoU and HA and follows the Association and Tracks Management Module \ref{Tracking Management Module}.
Hence, the successfully associated trajectories are denoted as $m\mathcal{T}^t$ and update their states by Eq.\ref{eq:upd_x},\ref{eq:upd_P} with the successfully associated detections $m\mathcal{G}^t$. The unsuccessfully associated tracks $u\mathcal{T}^t$ retain their previous state without update, and the unsuccessfully associated detections $u\mathcal{G}^t$ initialize new tracks. At time step $t+1$, all active tracks are predicted by Eq.(\ref{eq:prediction_sta}, \ref{eq:prediction_P}). The \textbf{AOS Graph Lap CoMOT} serves as the first step in deriving the proposed approach of the next subsection.
\begin{figure}[htbp]
\centering
 \includegraphics[scale=0.45]{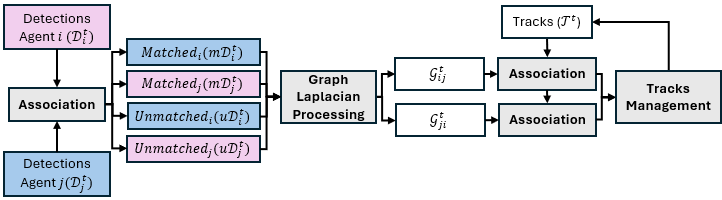}
  \caption{\textbf{TSA Graph Lap-CoMOT} of the Graph Lap-CoMOT approach}
  \label{fig:method1}
  \label{fig:method2}
\end{figure}

\subsection{Two Stages Association (TSA) Graph Lap-CoMOT}
\label{Method 2}
The \textbf{Two Stages Association (TSA) Graph Lap-CoMOT technique} (Fig.\ref{fig:method2}) is an advanced extension of the \textbf{AOS Graph Lap-CoMOT}. This method designs two stage association modules to capture unmatched trajectories from the first stage and combine obtained detections from different agents in slightly modified definition of anchors vector. Similarly to the AOS technique, the detections of the two CAVs are associated in order to provide additional information on the overlapped bounding boxes. Different anchors vectors are formulated based on the number of agents. Therefore, in case of two CAVs, two anchors vectors are defined for each spatial attribute along $x,y$, and $z$, and demonstrated for $x$ as:
\begin{align}
    \boldsymbol{a_{ij}^{t}} &=
    \begin{bmatrix}
     \boldsymbol{mx_{j,m}} & \boldsymbol{mx_{j,m}}& \boldsymbol{ux_{i,m}} & \boldsymbol{ux_{j,m}}
    \end{bmatrix}^T
    \label{eq:anchorsTAS1} \\
    \boldsymbol{a_{ji}^{t}} &=
    \begin{bmatrix}
     \boldsymbol{mx_{i,m}} & \boldsymbol{mx_{i,m}} & \boldsymbol{ux_{i,m}} & \boldsymbol{ux_{j,m}}
    \end{bmatrix}^T
    \label{eq:anchorsTAS2}
\end{align}
Therefore, different least-squares solutions of Eq.\ref{eq:LSMsol} are calculated. The fused and refined detections corresponding to the anchors vector $\boldsymbol{a_{ij}^t}$ are represented as $\boldsymbol{G_{ij}^t} \in \mathbb{R}^{N^t}$ and similarly to the $\boldsymbol{a_{ji}^t}$ as  $\boldsymbol{G_{ji}^t} \in \mathbb{R}^{N^t}$. Similar equations are followed for the $y,z$ attributes. Hence, the detections with all detections' attributes are $\mathcal{G}_{ij}^t \in \mathbb{R}^{N^t \times 7}$ and $\mathcal{G}_{ji}^t \in \mathbb{R}^{N^t \times 7}$. Firstly, $\mathcal{G}_{ij}^t$ are associated with the existing tracks $\mathcal{T}^t$ as described in the Association module of \ref{Tracking Management Module}. Upon successful associations, $m\mathcal{T}^t$ tracks update their state through the refined associated detection $m\mathcal{G}_{ij}^t$ by \ref{eq:upd_x},\ref{eq:upd_P}. In case of unmatched refined detections $u\mathcal{G}_{ij}^t$, new tracks are initialized. Additionally, the unmatched tracks $u\mathcal{T}^t$ are not discarded by the Tracks Management Module, instead are associated with the $\mathcal{G}_{ji}^t$ refined detections. Thereafter, the Association and Tracks Management Module \ref{Tracking Management Module} takes place with $m\mathcal{T}^t, m\mathcal{G}_{ji}^t$, the matched tracks, optimized detections and $ u\mathcal{T}^t$ and $u\mathcal{G}_{ji}^t$ the unmatched tracks and unmatched detections. This process continues based on the number of agents. At the next time step $t+1$, all active tracks are predicted by Eq.({\ref{eq:prediction_sta}},{\ref{eq:prediction_P}}). Regarding computational efficiency, Graph Laplacian Processing is based on an efficient sparse least-squares optimization framework, with complexity equal or lower than $\mathcal{O}((2N^t N^t)^2)$ (where $2N^t$ and $N^t$ the number of rows and columns of the corresponding Laplacian matrix, indicating actually the size of the topology) \cite{piperigkos2024graph}. Both \textbf{Graph Lap-CoMOT} approaches are summarized on \textbf{ Algorithm \ref{Graph Lap-CoMOT}}.

\begin{algorithm}
\small 
Input: Detections $\mathcal{D}_i^t$ from agent $i$ , Detections $\mathcal{D}_j^t$ from agent $j$, Tracks $\mathcal{T}^t$ \;
Output: Tracks $\mathcal{T}^{t+1}$ \;
\For {$t = 1, 2, \ldots, ...T$}{
For each spatial attribute $x,y,z$ associate 3D detections from different agents:
$m\mathcal{D}_i^t,m\mathcal{D}_j^t, u\mathcal{D}_i^t,u\mathcal{D}_j^t$=Associate\{$\mathcal{D}_i^t,\mathcal{D}_j^t$\} \;
\For {\textbf{AOS Graph Lap-CoMOT}}{
Calculate anchors vector $\boldsymbol{a^t}$ by Eq.\ref{eq:anchorsAOS} and refined-fused set of detections $\mathcal{G}^t$\;
\textbf{One Association Stage}\;
Associate $\mathcal{G}^t$ with existing tracks via Association module \ref{Tracking Management Module}:
$m\mathcal{T}^t, m\mathcal{G}^t, u\mathcal{T}^t,u\mathcal{G}^t$=Associate\{$\mathcal{G}^t,\mathcal{T}^t$\}\;
 Update, Initialize, Terminate via Tracks Management module \ref{Tracking Management Module}\;
}
\For {\textbf{TSA Graph Lap-CoMOT}}{ 
Calculate anchors vectors $\boldsymbol{a_{ij}^t}$, $\boldsymbol{a_{ji}^t}$ by Eq.\ref{eq:anchorsTAS1}, \ref{eq:anchorsTAS2} and refined-fused set of detections $\mathcal{G}_{ij}^t, \mathcal{G}_{ji}^t$\;
\textbf{First Association Stage}\; 
Associate $\mathcal{G}_{ij}^t$ with existing tracks via Association module \ref{Tracking Management Module}:
$m\mathcal{T}^t, m\mathcal{G}_{ij}^t, u\mathcal{T}^t,u\mathcal{G}_{ij}^t$=Associate\{$\mathcal{G}_{ij}^t,\mathcal{T}^t$\} \;
Update, Initialize via Association module \ref{Tracking Management Module}\;
\textbf{Second Association Stage}\; 
Associate $\mathcal{G}_{ji}^t$ with unmatched trajectories via Association module of \ref{Tracking Management Module}:
$m\mathcal{T}^t, m\mathcal{G}_{ji}^t, u\mathcal{T}^t,u\mathcal{G}_{ji}^t$=Associate\{$\mathcal{G}_{ji}^t,u\mathcal{T}^t$\} \;
Update, Initialize, Terminate via Association and Tracks Management modules \ref{Tracking Management Module}\;
}
Predict Tracks for  t+1 by  Eq.\ref{eq:prediction_sta}, \ref{eq:prediction_P} \;
}
\caption{\textbf{Graph Lap-CoMOT framework}}
\label{Graph Lap-CoMOT}
\end{algorithm}

\begin{table*}
\centering
\caption{Average Tracking Results in V2V4Real \cite{10203124}. Plus (minus) sign indicate the rate of accuracy improvement (decrease) with respect to the maximum deviation of the state-of-the-art CoMOT methods.}  
\resizebox{10cm}{!}{
\begin{tabular}{|c|c|c|c|c|c|}
\hline
Method & $AMOTA \ (\%)$ ($\uparrow$)  & $AMOTP \ (\%)$ ($\uparrow$) & $sAMOTA \ (\%)$ ($\uparrow$)  & $MT \ (\%)$ ($\uparrow$)\\
\hline
\textbf{V2V4Real $+$ CoBEV} & 36.18 &  53.14 & 75.2& 59.39\\
\textbf{DMSTrack}  &42.14& 57.05 & \textbf{84.02} & 65.07\\
\textbf{TSA Graph Lap-CoMOT} &
\textbf{42.44 (+17.3\%)} &  \textbf{61.64 (+15.99\%)}& 83.62 (-0.47\%) & \textbf{71.16 (+19.82\%)}\\
\hline
\end{tabular}
}
\label{tab:evaluate_sequences}
\end{table*}
\begin{table}[ht]
\centering
\caption{Ablation Study on three indicative V2V4Real Sequences. Plus (minus) sign indicate the rate of accuracy improvement (decrease) with respect to the DMSTrack \cite{chiu2024probabilistic}}  
\resizebox{9cm}{!}{
\begin{tabular}{|c|c|c|c|c|c|}
\hline
Sequence & Method & $AMOTA \ (\%)$  & $AMOTP \ (\%)$ & $sAMOTA \ (\%)$ & $MT \ (\%)$ \\
\hline
&\textbf{DMSTrack}  &48.06 & 60.46  & 81.95 & 60\\
Sequence 0000  &\textbf{AOS Graph Lap-CoMOT} &  49.34 (+2.66\%) &  68.27 (+12.92\%) & 87.81 (+7.15\%) &70 (+16.67\%)\\       
&\textbf{TSA Graph Lap-CoMOT} &
\textbf{54.63 (+13.67\%)} &  \textbf{71.11 (+17.61\%)}  & \textbf{91.43 (+11.56\%)} &\textbf{80 (+33.33\%)}\\
\hline
&\textbf{DMSTrack}  &  39.42 &50.18&68.22&42.86\\
Sequence 0002  &\textbf{AOS Graph Lap-CoMOT} &
44.56 (+13.04\%)&57.53 (+14.65\%)&83.93 (+23.03\%)&42.86\\         
&\textbf{TSA Graph Lap-CoMOT} &  \textbf{46.64 (+18.31\%)}&\textbf{60.13 (+19.82\%)}&\textbf{86.26 (+26.44\%)}&\textbf{42.86}\\
\hline        
&\textbf{DMSTrack}  &  46.75&62.40&91.16&93.33\\
Sequence 0007  &\textbf{AOS Graph Lap-CoMOT} &46.94 (+0.41\%)&63.56 (+1.86\%) &\textbf{91.68 (+0.57\%)}&93.33\\ 
&\textbf{TSA Graph Lap-CoMOT} &  \textbf{47.98 (+2.63\%)}&\textbf{67.80 (+8.65\%)}&89.74 (-1.56\%)&\textbf{96.67 (+3.58\%)}\\  
\hline
\end{tabular}
}
\label{tab:Abla}
\end{table}

\section{Numerical Results}
\subsection{Experimental Setup and Metrics}
Experiments have been conducted in the well-known V2V4Real \cite{10203124} dataset on an NVIDIA RTX 4090 GPU. The dataset training split includes 32 driving sequences and the testing 9 driving sequences from two simultaneously driven vehicles. The frame rate is 10Hz. 
Our proposed method ensures real-time feasibility by sharing only the bounding box parameters among agents, and thus, agent $j$ with $M_j$ bounding boxes, has to transmit in total $M_j \times{7}$ floats, since 7 are the parameters describing each bounding box, ensuring low latency during the communication.
Additionally, we define the tracking parameters $age$ = 2, and $hits$ = 3 to balance robustness and adaptability. We have tested our framework in the testing sequences and we compared our method with the state-of-the-art V2V4Real and the DMSTrack \cite{chiu2024probabilistic}, a deep-learning CoMOT approach.
The \textbf{Graph Lap-CoMOT} and DMSTrack approaches utilize PointPillar detector \cite{8954311} with 55\% detection accuracy. DMSTrack sequentially associates the 3D detections of the ego vehicle to the tracks, and the unmatched tracks with the bounding boxes of the other. Also, induces noise covariance matrix for each detection which is calculated by a DNN. Furthermore, we employed the V2V4Real method using multi-agent CoBEV Detector \cite{xu2022cobevt} with higher detection accuracy (66.5\%) than PointPillar for challenging evaluation.
We employ the evaluation metrics in 3D MOT of \cite{9341164} including 1) Average Multi-Object Tracking Accuracy (AMOTA), counting the errors (False Positives (FP), False Negatives (FN), Identity Switches(IDSW)) with respect to Ground Truth (GT), 2) Average Multi-Object Tracking Precision (AMOTP) calculating the overlap between the predicted tracked objects with the Ground Truth objects with respect to the True Positives, 3) scaled AMOTA (sAMOTA), a linearized AMOTA across various confident thresholds, 4) Mostly Tracked (MT), the percentage of the correct tracking of objects over 80\% of their life, 6) Multi-Object Tracking Precision (MOTP) calculating the overlap between the predicted tracked objects and GT with respect to the TP with 0.25 overlapping threshold.

\subsection{Evaluation Study}
Table \ref{tab:evaluate_sequences} demonstrates the performance of the \textbf{TSA Graph Lap-CoMOT} method and the two baseline methods on average across all testing sequences from the V2V4Real dataset. The \textbf{TSA Graph Lap-CoMOT} consistently achieves superior performance across the average sequences, with the maximum \textbf{17.3\%} improvement in AMOTA. Hence, the two stages association of tracks with refined detections captures potential occluded objects increasing the TP, reducing FN and thus, performs high tracking accuracy. 
Furthermore, the \textbf{TSA Graph Lap-CoMOT} reduces the noise in detections using the Graph Laplacian Operator within a fully connected graph via least-square minimization, and outperforms in tracking precision with \textbf{15.99\%} improvement in AMOTP.
Moreover, it enhances MT by \textbf{19.82\%} by tracking most objects for over of 80\% of their lifetime. However, \textbf{TSA Graph Lap-CoMOT} performs 0.47\% below the DMSTrack in sAMOTA. 
Therefore, \textbf{TSA Graph Lap-CoMOT} achieves superior performance across most tracking metrics by fusing the centroids of detections and reducing positional error via the Graph Laplacian operator, while captures unmatched trajectories preventing false termination with the two stages association.

To further enrich our ablation study, Table \ref{tab:Abla} demonstrates the tracking performance of both \textbf{Graph Lap-CoMOT} and DMSTrack methods on three indicative V2V4Real testing sequences, highlighting the benefits of two association stages. The \textbf{TSA Graph Lap-CoMOT} achieves superior performance with maximum improvements of \textbf{18.31\%}, \textbf{19.82\%}, \textbf{26.44\%}, \textbf{33.33\%} in AMOTA, AMOTP, sAMOTA, MT respectively over DMSTrack.
Additionally, \textbf{TSA Graph Lap-CoMOT} outperforms with maximum improvements up to \textbf{10.72\%}, \textbf{6.67\%}, \textbf{4.12\%}, \textbf{14.29\%} in all evaluated metrics respectively, over the one association stage, \textbf{AOS Graph Lap-CoMOT}. 
This demonstrates its effectiveness in addressing unmatched trajectories by preventing premature termination via the two association stages. Moreover, the \textbf{AOS Graph Lap-CoMOT} accomplishes significant improvements across various metrics despite being a simplified and single stage association method of the \textbf{TSA Graph Lap-CoMOT}. Specifically, it achieves a \textbf{13.04\%} improvement in AMOTA and \textbf{23.03\%} in sAMOTA for Sequence 0002, with more TP and consequently fewer FN tracked objects revealing and exploiting spatial geometric structures on the fully connected graph instead of DMSTrack. Additionally, a \textbf{14.65\%} in AMOTP and a \textbf{16.67\%} in MT improvement for Sequences 0002 and 0000 demonstrate the reduced spatial error of multi-agent detections via the Graph Laplacian Operator. 
Therefore, \textbf{TSA Graph Lap-CoMOT} exploits spatial coherences between multi-agent detections formulating a fully connected graph topology while addresses potential occluded objects with the two association stages.

\begin{figure}[htbp]
 \centering
 \includegraphics[scale=0.5]{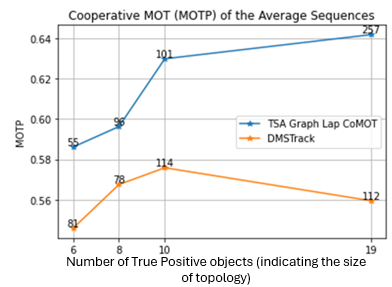}
  \caption{Impact of increasing graph topology size on MOTP on average. The $x$ and $y$-axis demonstrate the number of True Positive (TP) objects per frame and average MOTP, respectively. Bullet points indicate the frequency of each TP count per frame. \textbf{TSA Graph Lap-CoMOT} enhances tracking performance as the graph topology size increases.}
 \label{fig:detections_vsmotp}
\end{figure}

Fig.\ref{fig:detections_vsmotp} highlights the impact on tracking precision when the size of graph topology increases, demonstrating in fact the scalability of \textbf{TSA Graph Lap-CoMOT}. Higher localization accuracy (MOTP) is achieved on average through \textbf{TSA Graph Lap CoMOT} when 6 TP objects appear across the testing sequences, despite the higher frequency of appearance in DMSTrack. Moreover, a maximum MOTP up to \textbf{64.55\%} is achieved once again by our approach, when 19 TP objects exist across the testing sequences. 
Therefore, this fact emphasizes the significance of Graph Laplacian Processing in fusing diverse and growing number of bounding boxes. 

\begin{figure}[htbp]
  \centering
  \subfloat[Precise localization of the object by the TSA Graph Lap-CoMOT, as indicated by the red arrow, Frame 91]{\includegraphics[width=0.7\linewidth]{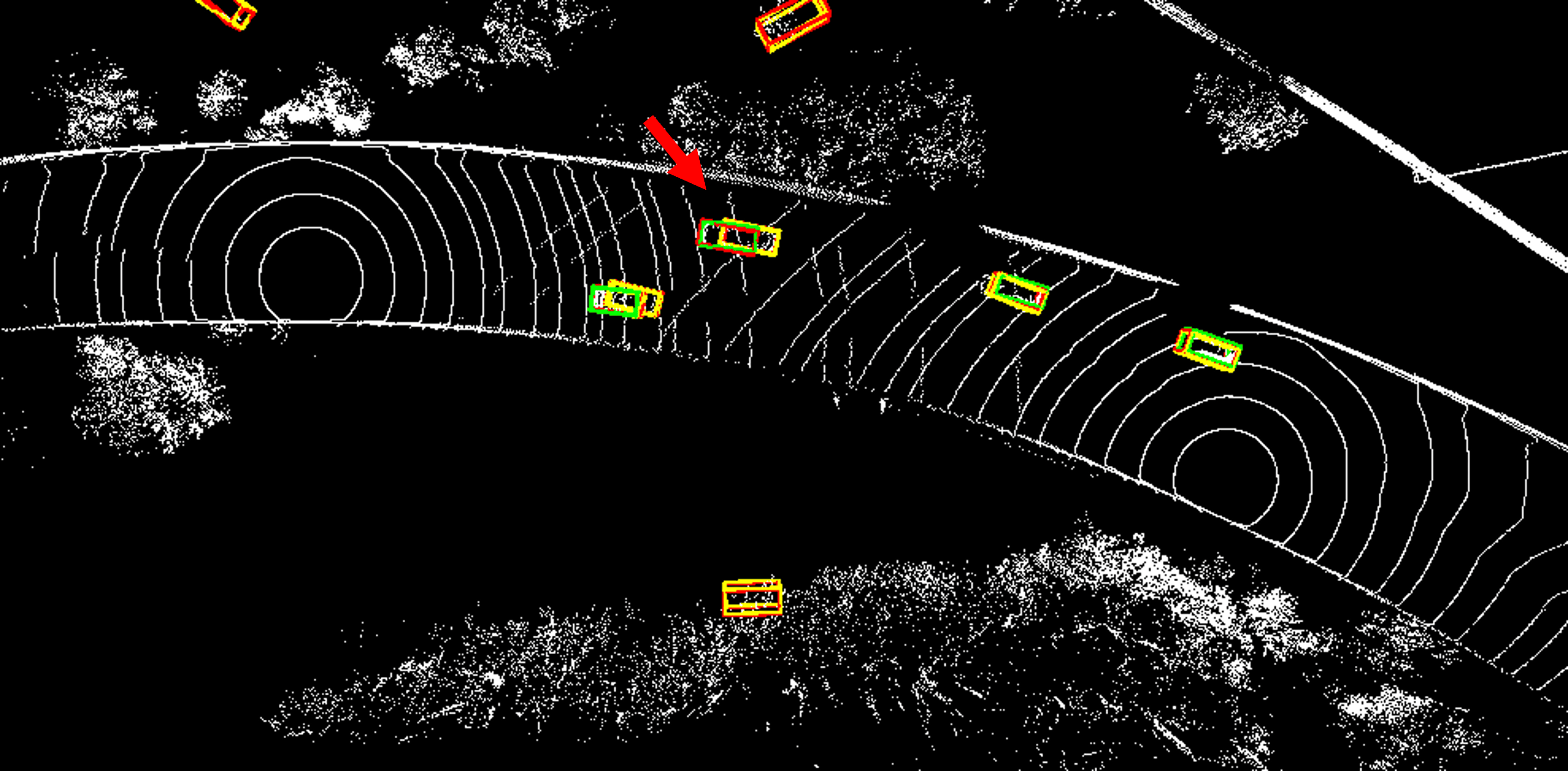}} \\
  \subfloat[Object capturing and precise localization, while the baseline fails, Frame 74]{\includegraphics[width=0.7\linewidth]{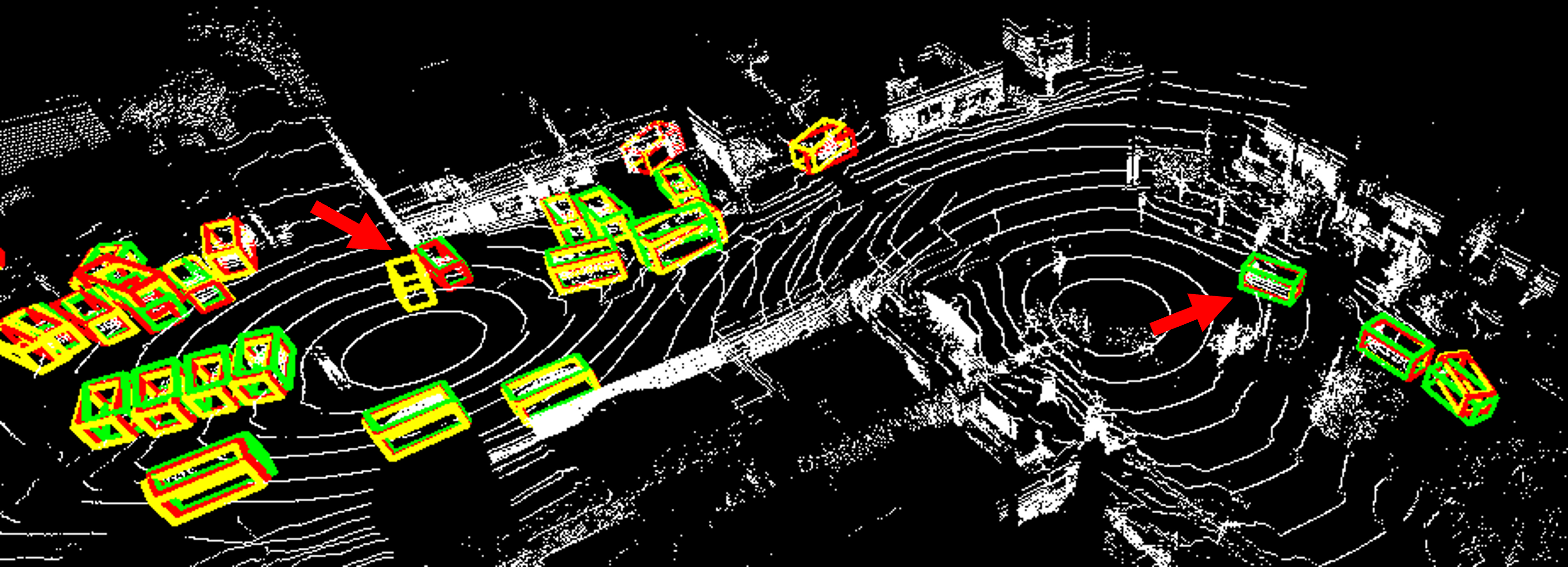}}
  \caption{Qualitative results of CoMOT based on assessing tracking accuracy and precision with GT (green), \textbf{TSA Graph Lap-CoMOT} (red) and DMSTrack (yellow).}
  \label{fig:qualitatitive}
\end{figure}

Finally, Fig.\ref{fig:qualitatitive} demonstrates GT, \textbf{TSA Graph Lap-CoMOT} and DMSTrack trajectories at frames 91 and 130 in Sequences 0000 and 0007, with green, red and yellow colors. In all cases, we can clearly observe that the proposed \textbf{TSA Graph Lap-CoMOT} significantly improved the location of each bounding box with respect to DMSTrack, as well as captured an object that DMSTrack failed to track succesfully. Therefore, \textbf{TSA Graph Lap-CoMOT} achieved competitive, accurate and precise tracking results in the real-world V2V4Real dataset.  

\section{Conclusion}
\label{conclusion}
In this paper, a novel CoMOT method is proposed to refine multi-agent multi-object detections by formulating a fully connected graph topology of individual bounding boxes that exploits the coherent diversity of detections using the Graph Laplacian framework. Spatial coherences on obtained detections are unveiled and 3D centroids of bounding boxes are optimized in a linear least-squares sense by reducing positional noise using both estimated differential coordinates and anchors. Thereafter, the refined detections update tracks' states via Kalman Filter,  achieving precise localization and more accurate identification of objects in the 3D space. To sum up, our \textbf{TSA Graph Lap-CoMOT} addresses the whole number of detections from the cooperative agents in a unified manner, fusing multi-agent information via Graph Laplacian Processing technique and outperforming state-of-the-art CoMOT methods. 
Future work will involve fusing multi-modal sensor data (e.g., LiDAR and Camera) from 
multiple agents and testing our concept to other related real-world and synthetic datasets.

\begin{thebibliography}{10}
\providecommand{\url}[1]{#1}
\csname url@samestyle\endcsname
\providecommand{\newblock}{\relax}
\providecommand{\bibinfo}[2]{#2}
\providecommand{\BIBentrySTDinterwordspacing}{\spaceskip=0pt\relax}
\providecommand{\BIBentryALTinterwordstretchfactor}{4}
\providecommand{\BIBentryALTinterwordspacing}{\spaceskip=\fontdimen2\font plus
\BIBentryALTinterwordstretchfactor\fontdimen3\font minus \fontdimen4\font\relax}
\providecommand{\BIBforeignlanguage}[2]{{%
\expandafter\ifx\csname l@#1\endcsname\relax
\typeout{** WARNING: IEEEtran.bst: No hyphenation pattern has been}%
\typeout{** loaded for the language `#1'. Using the pattern for}%
\typeout{** the default language instead.}%
\else
\language=\csname l@#1\endcsname
\fi
#2}}
\providecommand{\BIBdecl}{\relax}
\BIBdecl

\bibitem{10608725}
L.~Alfeqy, \textit{et al.}, ``Bevsort: Bird eye view lidar multi object tracking,'' in \emph{2024 IEEE 22nd Mediterranean Electrotechnical Conference (MELECON)}, 2024, pp. 7--12.

\bibitem{10204123}
Y.~Chen, \textit{et al.}, ``Voxelnext: Fully sparse voxelnet for 3d object detection and tracking,'' in \emph{2023 IEEE/CVF Conference on Computer Vision and Pattern Recognition (CVPR)}, 2023, pp. 21\,674--21\,683.

\bibitem{9352500}
H.~Wu, \textit{et al.}, ``3d multi-object tracking in point clouds based on prediction confidence-guided data association,'' \emph{IEEE Transactions on Intelligent Transportation Systems}, vol.~23, no.~6, pp. 5668--5677, 2022.

\bibitem{Yin_2021_CVPR}
T.~Yin, \textit{et al.}, ``Center-based 3d object detection and tracking,'' in \emph{Proceedings of the IEEE/CVF Conference on Computer Vision and Pattern Recognition (CVPR)}, June 2021, pp. 11\,784--11\,793.

\bibitem{9341164}
X.~Weng, \textit{et al.}, ``3d multi-object tracking: A baseline and new evaluation metrics,'' in \emph{2020 IEEE/RSJ International Conference on Intelligent Robots and Systems (IROS)}, 2020, pp. 10\,359--10\,366.

\bibitem{9537608}
N.~Piperigkos, \textit{et al.}, ``Graph laplacian diffusion localization of connected and automated vehicles,'' \emph{IEEE Transactions on Intelligent Transportation Systems}, vol.~23, no.~8, pp. 12\,176--12\,190, 2022.

\bibitem{chiu2024probabilistic}
H.-K. Chiu, \textit{et al.}, ``Probabilistic 3d multi-object cooperative tracking for autonomous driving via differentiable multi-sensor kalman filter,'' in \emph{2024 IEEE International Conference on Robotics and Automation (ICRA)}.\hskip 1em plus 0.5em minus 0.4em\relax IEEE, 2024, pp. 18\,458--18\,464.

\bibitem{10587163}
H.~Nguyen, \textit{et al.}, ``Multi-objective multi-agent planning for discovering and tracking multiple mobile objects,'' \emph{IEEE Transactions on Signal Processing}, vol.~72, pp. 3669--3685, 2024.

\bibitem{10203124}
R.~Xu, \textit{et al.}, ``V2v4real: A real-world large-scale dataset for vehicle-to-vehicle cooperative perception,'' in \emph{2023 IEEE/CVF Conference on Computer Vision and Pattern Recognition (CVPR)}, 2023, pp. 13\,712--13\,722.

\bibitem{9812038}
R.~Xu, et~al., ``Opv2v: An open benchmark dataset and fusion pipeline for perception with vehicle-to-vehicle communication,'' in \emph{2022 International Conference on Robotics and Automation (ICRA)}, 2022, pp. 2583--2589.

\bibitem{8653482}
M.~Jiang, \textit{et al.}, ``Multi-agent deep reinforcement learning for multi-object tracker,'' \emph{IEEE Access}, vol.~7, pp. 32\,400--32\,407, 2019.

\bibitem{10430224}
S.~Su, \textit{et al.}, ``Collaborative multi-object tracking with conformal uncertainty propagation,'' \emph{IEEE Robotics and Automation Letters}, vol.~9, no.~4, pp. 3323--3330, 2024.

\bibitem{10588576}
H.~Su, \textit{et al.}, ``Cooperative 3d multi-object tracking for connected and automated vehicles with complementary data association,'' in \emph{2024 IEEE Intelligent Vehicles Symposium (IV)}, 2024, pp. 285--291.

\bibitem{10148929}
Z.~Meng, \textit{et al.} and Xia, ``Hydro-3d: Hybrid object detection and tracking for cooperative perception using 3d lidar,'' \emph{IEEE Transactions on Intelligent Vehicles}, vol.~8, no.~8, pp. 4069--4080, 2023.

\bibitem{9711021}
C.~Luo, \textit{et al.}, ``Exploring simple 3d multi-object tracking for autonomous driving,'' in \emph{2021 IEEE/CVF International Conference on Computer Vision (ICCV)}, 2021, pp. 10\,468--10\,477.

\bibitem{sorkine2005laplacian}
O.~Sorkine, ``Laplacian mesh processing,'' \emph{Eurographics (State of the Art Reports)}, vol.~4, no.~4, p.~1, 2005.

\bibitem{piperigkos2024graph}
N.~Piperigkos, \textit{et al.}, ``Graph laplacian processing based multi-modal localization backend for robots and autonomous systems,'' \emph{IEEE Transactions on Cognitive and Developmental Systems}, 2024.

\bibitem{8954311}
A.~Lang, \textit{et al.}, ``Pointpillars: Fast encoders for object detection from point clouds,'' in \emph{2019 IEEE/CVF Conference on Computer Vision and Pattern Recognition (CVPR)}, 2019, pp. 12\,689--12\,697.

\bibitem{xu2022cobevt}
R.~Xu, \textit{et al.}, ``Cobevt: Cooperative bird's eye view semantic segmentation with sparse transformers,'' \emph{arXiv preprint arXiv:2207.02202}, 2022.

\end{thebibliography}

\end{document}